\documentclass{article}
\usepackage{ijcai20}

\usepackage{times}
\usepackage{soul}
\usepackage{url}
\usepackage[hidelinks]{hyperref}
\usepackage[utf8]{inputenc}
\usepackage[small]{caption}
\usepackage{graphicx}
\usepackage{amsmath}
\usepackage{amsthm}
\usepackage{booktabs}
\usepackage{courier}  
\usepackage{graphicx}  
\usepackage{multicol}
\usepackage{multirow}
\usepackage{graphicx, multicol, latexsym, amsmath, amssymb}
\usepackage{mathrsfs}
\usepackage{amssymb}
\newcommand{\argmin}{\operatornamewithlimits{argmin}}

\usepackage{amsmath}
\usepackage{caption}
\usepackage{subcaption}
\usepackage{graphicx}
\usepackage[ruled,linesnumbered,lined]{algorithm2e}
\title{Decorrelated Clustering with Data Selection Bias}

\author{
Xiao Wang$^1$\and
Shaohua Fan$^1$\and
Kun Kuang$^2$\and
Chuan Shi$^1$\footnote{Corresponding Author.}\and
Jiawei Liu$^1$\And
Bai Wang$^1$\\
\affiliations
$^1$Beijing University of Posts and Telecommunications\\
$^2$Zhejiang University\\
\emails
xiaowang@bupt.edu.cn,
fanshaohua92@163.com,
kunkuang@zju.edu.cn,\\
\{shichuan, liu\_jiawei, wangbai\}@bupt.edu.cn
}

\begin{document}

\maketitle

\begin{abstract}

Most of existing clustering algorithms are proposed without considering the selection bias in data. In many real applications, however, one cannot guarantee the data is unbiased. Selection bias might bring the unexpected correlation between features and ignoring those unexpected correlations will hurt the performance of clustering algorithms. Therefore, how to remove those unexpected correlations induced by selection bias is extremely important yet largely unexplored for clustering. In this paper, we propose a novel Decorrelation regularized $K$-Means algorithm (DCKM) for clustering with data selection bias. Specifically, the decorrelation regularizer aims to learn the global sample weights which are capable of balancing the sample distribution, so as to remove unexpected correlations among features. Meanwhile, the learned weights are combined with $k$-means, which makes the reweighted $k$-means cluster on the inherent data distribution without unexpected correlation influence. Moreover, we derive the updating rules to effectively infer the parameters in DCKM. Extensive experiments results on real world datasets well demonstrate that our DCKM algorithm achieves significant performance gains, indicating the necessity of removing unexpected feature correlations induced by selection bias when clustering.
\end{abstract}

\section{Introduction}




One common hypothesis in traditional machine learning is that the data is drawn from an unbiased distribution, in which there are weak correlations between features~\cite{heckman1979sample,huang2007correcting}. However, in many real world applications, we cannot fully control the data  gathering process and always suffer from the data selection bias issue, which will inevitably cause the correlations between features.
\begin{figure}[ht]
\centering
\includegraphics[width=8cm]{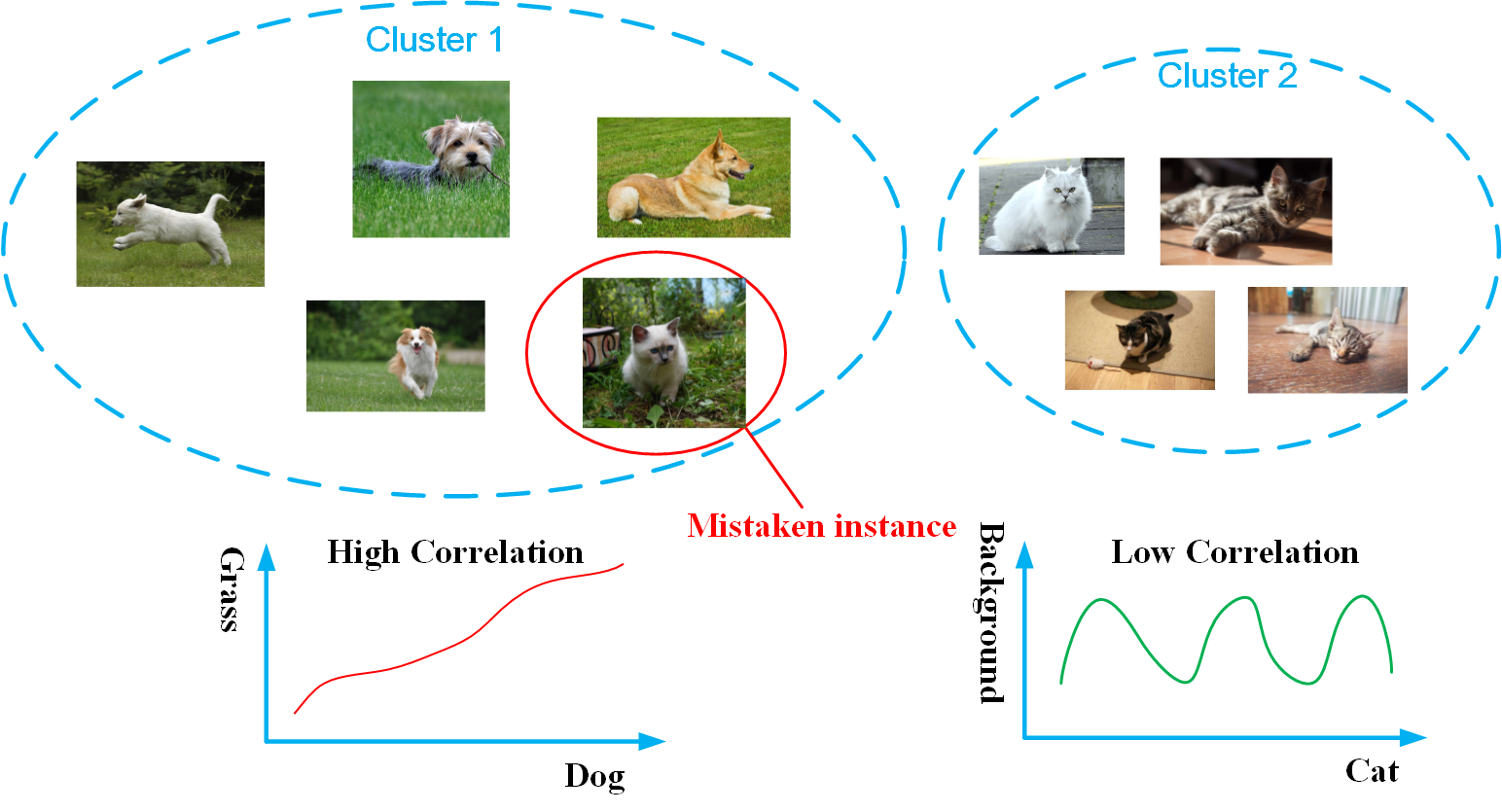} 
\caption{An example of clustering on data with high correlated features.}
\label{fig:fig1}
\end{figure}
Unexpected high feature correlation is undesirable, as it not only brings redundancy in features, but also causes the algorithm to unsatisfied results~\cite{zhang2018removing}.
Some literatures have studied the problem of removing the feature correlation effect in machine learning model~\cite{bengio2009slow,cogswell2015reducing,rodriguez2016regularizing,zhang2018removing}. They mainly focus on removing the feature correlation effect in neural networks by designing decorrelation components, which bring great benefits for representation learning.

\par Despite the enormous success of decorrelation in neural networks, the effect of data selection bias is severely underestimated in unsupervised learning scenario. Typically, clustering also suffers from the data selection bias issue~\cite{kriegel2009clustering}. Data selection bias may cause spurious correlation between features. Assuming one meaningless feature is mistakenly identified to correlate with one important feature, because of the presence of spurious correlation, the effect of this meaningless feature will be unconsciously strengthened, rendering the inherent data distribution unrevealed. Thus clustering on these data will inevitably result in poor performance. As depicted in Figure~\ref{fig:fig1}, given an image dataset with many dogs on the grass and some cats in various backgrounds, it is easy to draw a conclusion that grass features are highly correlated with dog features and cat features have low correlation with background. Therefore, when performing clustering algorithm on such biased dataset, any object on the grass, even a cat, will be clustered to the dog cluster with large probability. This implies that clustering is very easily misled by the presence of spurious correlations between features. However, most of existing clustering algorithm~\cite{hartigan1979algorithm,bachem2018scalable,schmidt2018fair,von2007tutorial} do not take the data selection bias into consideration, and the feature correlation effect in clustering is largely ignored.


\par Although it is promising to marry feature decorrelation with clustering, there are two unsolved challenges. (1) \textit{How to remove the correlations between features in high-dimensional scenarios}?  In real applications, the correlations between features might be very complex, especially in high-dimensional settings. Moreover, we have little prior knowledge about which correlations are unexpected and would hurt clustering performance. In practical, one possible way is to remove correlations between each targeted feature with the remaining features one by one, but obviously this method suffers from huge model complexity. Therefore, we need to design efficient feature decorrelation method. (2) \textit{How to make the feature decorrelation benefit for clustering}? Feature decorrelation and clustering are traditionally two independent tasks. Because they have different objectives, feature decorrelation does not necessarily lead to good clustering. Therefore, we need to discriminatively remove correlations for clustering. To achieve this goal, a task-oriented feature decorrelation framework is highly desirable. However, it is highly non-trivial to design a scalable feature decorrelation method for clustering problem, because feature decorrelation usually cannot be directly incorporated with clustering objective.



\par In this paper, we propose a novel Decorrelation regularized $K$-Means (DCKM) model for clustering on data with selection bias. Specifically, to decorrelate one targeted feature with the remaining features, a decorrelation regularizer is introduced to balance the remaining feature distributions through learning a global sample weight matrix. Meanwhile, the weight matrix is employed to reweight the $k$-mean loss. In this way, the weighted $k$-means and decorrelation regularizer are in a unified framework, causing that clustering results are not affected by unexpected correlated features. Moreover, we derive an effectively iterative updating rules to optimize the parameters of our model. Our contributions are summarized in the following three folds:
\begin{itemize}
	\item We investigate an important but seldom studied problem, i.e., clustering on data with selection bias. The correlation caused by the data selection bias is ubiquitous in real applications, while the effect of the correlation in clustering is largely unexplored.
									
	\item We propose a novel Decorrelation regularized $K$-Means (DCKM) model which removes the unexpected correlations among features for clustering by a decorrelation regularizer. Moreover, we derive an effectively updating algorithm to optimize the parameters of DCKM.
									
	\item We conduct comprehensive experiments, where the significant performance gains demonstrate the superiority of our method in clustering on the biased data.
\end{itemize}

\section{Preliminaries}

\paragraph{Notations.} In our paper, $n$ refers to the sample size, and $d$ is the dimensions of features. For a vector $\mathbf{v}\in\mathbb{R}^{d\times 1}$, $\mathbf{v}_i$ represents the $i$-th element of $\mathbf{v}$ and $||\mathbf{v}||^2_2=\sum^d_{i=1}\mathbf{v}_i^2$. For any matrix $\mathbf{X}\in\mathbb{R}^{n\times d}$, we denote $\mathbf{X}_{i.}$ and $\mathbf{X}_{.j}$ represent the $i$-th row and the $j$-th column in $\mathbf{X}$, respectively. And  $||\mathbf{X}||^2_F=\sum_{i=1}^n\sum_{j=1}^d \mathbf{X}_{ij}^2$.

\paragraph{Problem Definition. Clustering on Data with Selection Bias}.
Given $n$ samples with $d$-dimensional features, represented by $\mathbf{X}\in\mathbb{R}^{n\times d}$, the task is to learn a robust clustering model, which will not be affected by the unexpected correlations between features, to partition the $n$ samples into predefined $K$ disjoint clusters $\{C_1, \cdots, C_K\}$.

\paragraph{Definition 1. Remaining Features.} If we treat the $j$-th feature of $\mathbf{X}$ (i.e., $\mathbf{X}_{.j}$) as targeted feature, $\mathbf{X}_{.-j}=\mathbf{X}\setminus\mathbf{X}_{.j}$ are regarded as remaining features, which is from $\mathbf{X}$ by replacing its $j$-th column as 0.

\paragraph{Definition 2. Treated Group and Control Group.} Given the targeted feature $\mathbf{X}_{.j}$, if the $j$-th feature of sample $i$: $\mathbf{X}_{ij}=1$\footnote{Please note that, without losing any generality, here we assume all the features are binary for the ease of discussion and understanding~(categorical and continuous features can be converted to binary ones through binning and one-hot encoding).}, then the sample $\mathbf{X}_{i.}$ is a treated sample, and the treated group is a sample set $TG_j = \{\mathbf{X}_{i.}|\mathbf{X}_{ij}=1\}$; otherwise, the sample set $CG_j = \{\mathbf{X}_{i.}|\mathbf{X}_{ij}=0\}$ is a control group.

\par It is well recognized that $k$-means is one of the most representative clustering algorithms. Thus, to validate the necessity of decorrelation when clustering, we focus on $k$-means algorithm and propose a novel decorrelation regularized $k$-means method. Here, we first introduce some preliminaries in $k$-means clustering.

\paragraph{$K$-means and matrix factorization.} The classical $k$-means clustering is a centroid-based clustering method, which partitions the data space into a structure known as Voronoi diagram. Besides, the G-orthogonal non-negative matrix factorization (NMF) is equivalent to relaxed $k$-means clustering~\cite{ding2005nonnegative}, which can be reformulated as:
\begin{equation}
	\begin{aligned}
		&\min_{\mathbf{F}, \mathbf{G}}\sum^n_{i=1}||\mathbf{X}_{i.}-\mathbf{G}_{i.}\cdot\mathbf{F}^T||^2_2, \\
&s.t.\quad\mathbf{G}_{ik}\in\{0,1\}, \sum_{k=1}^K \mathbf{G}_{ik}=1, \forall i=1,2,\cdots,n,
	\end{aligned}
	\label{equ:km-data}
\end{equation}
where $\mathbf{F}\in\mathbb{R}^{d\times K}$ is the cluster centroid matrix,  $\mathbf{G}\in \mathbb{R}^{n\times K}$ is the cluster assignment matrix, each row of which satisfies the \textit{1-of-K} coding scheme, i.e., if data point $\mathbf{X}_{i.}$ is assigned to $k$-th cluster, then $\mathbf{G}_{ik}=1$; otherwise, $\mathbf{G}_{ik}=0$.


\section{Decorrelation Regularized $K$-means}

\subsection{Decorrelation Regularizer}
Recalling the example in Figure~\ref{fig:fig1}, we assume the $j$-th feature represents whether the image has dog feature and the $t$-th feature indicates whether the image has grass feature. If the majority of dogs are on the grass, then the $j$-th and the $t$-th feature will be highly correlated. As a result, when perform clustering on such data, the $t$-th feature, i.e., the grass feature, will probably mislead the algorithm to cluster other kinds of images with grass and dogs into the same cluster. One alternative solution to alleviate the data selection bias is to add extra dog images with other backgrounds, so that the $j$-th feature will not correlate with the $t$-th feature, but it is difficult to obtain extra data in many real applications.

Instead, we adjust data distribution by learning a sample weight for each sample so that all the features trend to be independent~\cite{shen2018causally,kuang2018stable,kuang2020stable,shen2020stable}. Specifically, we first focus on how to remove correlation between the $j$-th targeted feature $\mathbf{X}_{.j}$ and the correspondingly remaining features $\mathbf{X}_{.-j}$.

\paragraph{Single feature decorrelation regularizer.} If the targeted feature $\mathbf{X}_{.j}$ correlates with the remaining features $\mathbf{X}_{.-j}$, the treated and control groups, $TG_j$ and $CG_j$, will have different distributions on $\mathbf{X}_{.-j}$. Once we balance the distributions between $TG_j$ and $CG_j$, we are able to reduce the correlation between the targeted feature and the correspondingly remaining features. As moments can uniquely determine a distribution~\cite{shen2018causally}, we use the first-order moment to measure the distributions. Specifically, for the remaining features in treated group $TG_j$, the first-order moment is:

\begin{equation}
    \bar{\mathbf{X}}_{.-j}=\frac{\mathbf{X}^{T}_{.-j}\cdot\mathbf{X}_{.j}}{\mathbf{1}_n^T\cdot \mathbf{X}_{.j}},
	\label{equ:W}
\end{equation}
where $\mathbf{1}_n=[1,1,\cdots,1]\in\mathbb{R}^{n\times 1}$. Similarly, the first-order moment of the remaining feature in control group $CG_j$ is:

 \begin{equation}
    \hat{\mathbf{X}}_{.-j}=\frac{\mathbf{X}^{T}_{.-j}\cdot(\mathbf{1}_n-\mathbf{X}_{.j})}{\mathbf{1}_n^T\cdot (\mathbf{1}_n-\mathbf{X}_{.j})}.
	\label{equ:W}
\end{equation}

To balance the moments $\bar{\mathbf{X}}_{.-j}$ and $\hat{\mathbf{X}}_{.-j}$, we introduce the sample weights $\mathbf{w}^{j}\in\mathbb{R}^{n\times 1}$ to adjust the value of moments, which can be learned by:

\begin{equation}
    \begin{aligned}
    \mathbf{w}^{j}=&\argmin_{\mathbf{w}^{j}}||\frac{\mathbf{X}^{T}_{.-j}\cdot(\mathbf{w}^{j}\odot \mathbf{X}_{.j})}{{\mathbf{w}^{j}}^T\cdot \mathbf{X}_{.j}}\\
    &-\frac{\mathbf{X}^{T}_{.-j}\cdot(\mathbf{w}^{j}\odot (\mathbf{1}_n-\mathbf{X}_{.j}))}{{\mathbf{w}^{j}}^T\cdot (\mathbf{1}_n-\mathbf{X}_{.j})}||^2_2,
	\label{equ:Balance}
	\end{aligned}
\end{equation}
where `$\odot$' refers to the Hadamard product. The first term $\frac{\mathbf{X}^{T}_{.-j}\cdot(\mathbf{w}^j\odot \mathbf{X}_{.j})}{{\mathbf{w}^{j}}^T\cdot \mathbf{X}_{.j}}$ is the weighted moment of $TG_j$ and the second term $\frac{\mathbf{X}^{T}_{.-j}\cdot(\mathbf{w}^j\odot (\mathbf{1}_n-\mathbf{X}_{.j}))}{{\mathbf{w}^{j}}^T\cdot (\mathbf{1}_n-\mathbf{X}_{.j})}$ is the weighted moment of $CG_j$. By optimizing Eq.~(\ref{equ:Balance}), the two terms will be balanced. After remaining features balancing, the targeted feature selection bias will be corrected and the correlation between the targeted feature and remaining features will tend to be removed.

\paragraph{Global feature decorrelation regularizer.} Note that the above method is to remove the correlation between a single targeted feature $\mathbf{X}_{.j}$ with the remaining features $\mathbf{X}_{.-j}$. However, we need to remove the correlations of all features with the correspondingly remaining features. This implies that we need to learn $n\times d$ sample weights, which is apparently infeasible in high-dimensional scenarios. However, because $d$ sets of sample weights $\{\mathbf{w}^{j}\}_{j=1}^{d}$ are used to adjust the same set of $n$ samples, the sample weights for different targeted feature can be shared. Thus we introduce a global balancing method as the decorrelation regularizer. Specially, we add all the single feature remaining feature balancing term together, in which each balancing term is formulated by setting each feature as targeted feature, and for all the remaining feature balancing term, they use the same set of sample weights $\mathbf{w}\in\mathbb{R}^{n\times1}$:

\begin{equation}
    \sum^d_{j=1}||\frac{\mathbf{X}^{T}_{.-j}\cdot(\mathbf{w}\odot \mathbf{X}_{.j})}{\mathbf{w}^T\cdot \mathbf{X}_{.j}}-\frac{\mathbf{X}^{T}_{.-j}\cdot(\mathbf{w}\odot (\mathbf{1}_n-\mathbf{X}_{.j}))}{\mathbf{w}^T\cdot (\mathbf{1}_n-\mathbf{X}_{.j})}||^2_2.
	\label{equ:Balance_all}
\end{equation}

As we can see from Eq.~(\ref{equ:Balance_all}), the global sample weights $\mathbf{w}$ simultaneously balance all the remaining feature terms, which yields the correlations between all features tend to be removed.

\subsection{Decorrelation Regularized $K$-means}
\par In the traditional $k$-means model Eq.~(\ref{equ:km-data}), the cluster centroid $\mathbf{F}$ and the cluster assignment $\mathbf{G}$ are learned on the original feature $\mathbf{X}$. But the unexpected highly correlated features may confuse the data distribution, which yields to unsatisfied clustering results. Because the sample weights $\mathbf{w}$ learned from the decorrelation regularizer are capable of globally decorrelating the features, we propose to use the weights to reweight the $k$-means loss and jointly optimize the weighted $k$-means loss and decorrelation regularizer:

\begin{equation}
\begin{aligned}
&\min_{\mathbf{w},\mathbf{F},\mathbf{G}} \sum^n_{i=1}\mathbf{w}_i\cdot||\mathbf{X}_{i.}-\mathbf{G}_{i.}\cdot\mathbf{F}^T||^2_2,\\
&s.t.\quad\sum^{d}_{j=1}||\frac{\mathbf{X}^{T}_{.-j}\cdot(\mathbf{w}\odot \mathbf{X}_{.j})}{\mathbf{w}^T\cdot \mathbf{X}_{.j}}\\
&-\frac{\mathbf{X}^{T}_{.-j}\cdot(\mathbf{w}\odot (\mathbf{1}_n-\mathbf{X}_{.j}))}{\mathbf{w}^T\cdot (\mathbf{1}_n-\mathbf{X}_{.j})}||^2_2\leq\gamma_1,\\
&\qquad \mathbf{G}_{ik}\in\{0,1\},\sum^K_{k=1}\mathbf{G}_{ik}=1,\\ &\mathbf{w}\succeq0,||\mathbf{w}||^2_2\leq\gamma_2, (\sum^n_{i=1}\mathbf{w}_i-1)^2\leq\gamma_3.
\end{aligned}
\label{equ:model}
\end{equation}

The term $\mathbf{w}\succeq0$ constrains each of sample weights to be non-negative. With norm $||\mathbf{w}||^2_2\leq\gamma_2$, we can reduce variance of the sample weights to achieve stability. The formula $(\sum^n_{i=1}\mathbf{w}_i-1)^2\leq\gamma_3$ avoids all the sample weights to be 0.
\par Although DCKM still performs on data $\mathbf{X}$, the weight of each $\mathbf{X}_{i.}$ is no longer same. This weight adjusts the contribution of each data in the entire loss, so that the cluster centroid and the cluster assignment are learned on the decorrelated features which can better reveal real data distribution.

\subsection{Optimization}
The constrained matrix factorization objective Eq.~(\ref{equ:model}) is not convex, and we separate the optimization of Eq.~(\ref{equ:model}) into three subproblems and iteratively optimize them. Next we describe the optimization process in detail.

The function Eq.~(\ref{equ:model}) is equal to minimize $\mathcal{J}(\mathbf{w},\mathbf{F},\mathbf{G})$:
\begin{equation}
\begin{aligned}
\mathcal{J}(\mathbf{w},\mathbf{F}, \mathbf{G})&=||(\mathbf{X}-\mathbf{G}\cdot\mathbf{F}^T)\odot(\mathbf{w}\cdot \mathbf{1}_{d}^{T})^{1/2}||^2_F\\
&+\lambda_1\sum^{d}_{j=1}||\frac{\mathbf{X}^{T}_{.-j}\cdot(\mathbf{w}\odot \mathbf{X}_{.j})}{\mathbf{w}^T\cdot \mathbf{X}_{.j}}\\
&-\frac{\mathbf{X}^{T}_{.-j}\cdot(\mathbf{w}\odot (\mathbf{1}_n-\mathbf{X}_{.j}))}{\mathbf{w}^T\cdot (\mathbf{1}_n-\mathbf{X}_{.j})}||^2_2\\
&+\lambda_2||\mathbf{w}||^2_2+\lambda_3 (\sum^n_{i=1}\mathbf{w}_i-1)^2,\\
&s.t.\quad \mathbf{w}\succeq0, \mathbf{G}_{ik}\in\{0,1\},\sum^K_{k=1}\mathbf{G}_{ik}=1.
\end{aligned}
\label{equ:model_la}
\end{equation}

\par To optimize Eq.~(\ref{equ:model_la}), we iteratively update three parameters (i.e. $\mathbf{F}$, $\mathbf{G}$, $\mathbf{w}$), which are described below:

\paragraph{$\mathbf{F}$-subproblem}: When updating $\mathbf{F}$ with $\mathbf{w}$ and $\mathbf{G}$ in Eq.~(\ref{equ:model_la}) being fixed, we need to optimize the following objective function:

\begin{equation}
\begin{aligned}
\mathcal{J}(\mathbf{F})&=||(\mathbf{X}-\mathbf{G}\cdot\mathbf{F}^T)\odot(\mathbf{w}\cdot \mathbf{1}_{d}^{T})^{1/2}||^2_F,
\end{aligned}
\label{equ:model_J}
\end{equation}
which is a form of weighted $k$-means.
Taking derivative of $\mathcal{J}(\mathbf{F})$ with respect to $\mathbf{F}$, we get
\begin{equation}
\begin{aligned}
\frac{\partial \mathcal{J}(\mathbf{F})}{\partial \mathbf{F}}=&-2(\mathbf{X}^T\odot ({\mathbf{1}}_d\cdot \mathbf{w}^T))\cdot\mathbf{G}
\\&-2\mathbf{F}\cdot(\mathbf{G}^T\odot( {{\mathbf{1}}_K} \cdot {w}^{T}))\cdot\mathbf{G}.
\end{aligned}
\label{equ:model_F_de}
\end{equation}

Setting Eq.~(\ref{equ:model_F_de}) to 0, we can update $\mathbf{F}$ as:

\begin{equation}
\begin{aligned}
\mathbf{F}=(\mathbf{X}^T\odot (\mathbf{1}_d\cdot \mathbf{w}^T))\cdot\mathbf{G}\cdot((\mathbf{G}^T\odot( {{\mathbf{1}}_K} \cdot \mathbf{w}^{T}))\cdot\mathbf{G})^{-1}.
\end{aligned}
\label{equ:model_F_up}
\end{equation}

\paragraph{$\mathbf{G}$-subproblem}: When updating $\mathbf{G}$ with $\mathbf{F}$ and $\mathbf{w}$ in Eq.~(\ref{equ:model_la}) being fixed, we need to optimize the following objective function:

\begin{equation}
\begin{aligned}
\mathcal{J}(\mathbf{G})=\sum^n_{i=1}\mathbf{w}_i\cdot||\mathbf{X}_{i.}-\mathbf{G}_{i.}\cdot\mathbf{F}^T||^2_2,\\
s.t.\quad\mathbf{G}_{ik}\in\{0,1\}, \sum_{k=1}^K \mathbf{G}_{ik}=1.
\end{aligned}
\label{equ:model_J_G}
\end{equation}

We can solve Eq.~(\ref{equ:model_J_G}) by decoupling the data and assigning the cluster indicator for them one by one independently. In particular, we optimize $\mathbf{G}_{i.}$ for each sample $i$ respectively:
\begin{equation}
\begin{aligned}
&\min_{\mathbf{G}_{i.}}\mathbf{w}_i\cdot||\mathbf{X}_{i.}-\mathbf{G}_{i.}\cdot\mathbf{F}^T||^2_2,\\
&s.t.\quad \mathbf{G}_{ik}\in\{0,1\}, \sum_{k=1}^K \mathbf{G}_{ik}=1.
\end{aligned}
\label{equ:model_g}
\end{equation}
We can see that $\mathbf{w}_i$ will not influence the optimal $\mathbf{G}_{i.}$. Given the fact that $\mathbf{G}_{i.}$ satisfies \textit{1-of-K} coding scheme, there are $K$ candidates to be the solution of Eq.~(\ref{equ:model_g}), each of which is the $k$-th column of matrix $\mathbf{I}_K=[\mathbf{e}_1,\mathbf{e}_2,\cdots,\mathbf{e}_K]$. To be specific, we can perform an exhaustive search to find out the solution of Eq.~(\ref{equ:model_g}) as,
\begin{equation}
\mathbf{G}_{i.}^*=\mathbf{e}_k,
\label{equ:model_g1}
\end{equation}
where $k$ is decided as follows,
\begin{equation}
k = \argmin_j ||\mathbf{X}_{i.}-\mathbf{e}_j\cdot\mathbf{F}^T||.
\label{equ:model_g2}
\end{equation}

\paragraph{$\mathbf{w}$-subproblem}: When updating $\mathbf{w}$ with $\mathbf{F}$ and $\mathbf{G}$ in Eq.~(\ref{equ:model_la}) being fixed, we need to optimize the following objective function:

\begin{equation}
\begin{aligned}
\mathcal{J}(\mathbf{w})&=||(\mathbf{X}-\mathbf{G}\cdot\mathbf{F}^T)\odot(\mathbf{w}\cdot \mathbf{1}_{d}^{T})^{1/2}||^2_F\\
&+\lambda_1\sum^{d}_{j=1}||\frac{\mathbf{X}^{T}_{.-j}\cdot(\mathbf{w}\odot \mathbf{X}_{.j})}{\mathbf{w}^T\cdot \mathbf{X}_{.j}}\\
&-\frac{\mathbf{X}^{T}_{.-j}\cdot(\mathbf{w}\odot (\mathbf{1}_n-\mathbf{X}_{.j}))}{\mathbf{w}^T\cdot (\mathbf{1}_n-\mathbf{X}_{.j})}||^2_2\\
&+\lambda_2||\mathbf{w}||^2_2+\lambda_3(\sum^n_{i=1}\mathbf{w}_i-1)^2,\\
&s.t.\quad \mathbf{w}\succeq0.
\end{aligned}
\label{equ:model_W_16}
\end{equation}

We let $\mathbf{w}=\omega\odot\omega$ to ensure non-negativity of $\mathbf{w}$, where $\omega\in\mathbb{R}^{n\times 1}$. Then Eq.~(\ref{equ:model_W_16}) can be reformulated as:

\begin{equation}
\begin{aligned}
\mathcal{J}(\omega)&=||(\mathbf{X}-\mathbf{G}\cdot\mathbf{F}^T)\odot((\omega\odot\omega)\cdot \mathbf{1}_{d}^{T})^{1/2}||^2_F\\
&+\lambda_1\sum^{d}_{j=1}||\frac{\mathbf{X}^{T}_{.-j}\cdot(\omega\odot\omega\odot \mathbf{X}_{.j})}{\omega\odot\omega^T\cdot \mathbf{X}_{.j}}\\
&-\frac{\mathbf{X}^{T}_{.-j}\cdot(\omega\odot\omega\odot (\mathbf{1}_n-\mathbf{X}_{.j}))}{(\omega\odot\omega)^T\cdot (\mathbf{1}_n-\mathbf{X}_{.j})}||^2_2\\
&+\lambda_2||\omega\odot\omega||^2_2+\lambda_3(\sum^n_{i=1}\omega_i\odot\omega_i-1)^2.
\end{aligned}
\label{equ:model_W}
\end{equation}

The partial gradient of term $\mathcal{J}(\omega)$ with respect to $\omega$ is:

\begin{equation}
\begin{aligned}
\frac{\partial \mathcal{J}(\omega)}{\partial \omega}&=(\mathbf{1}_n^T\cdot((\mathbf{X}^T-\mathbf{F}\cdot\mathbf{G}^T)\odot(\mathbf{X}^T-\mathbf{F}\cdot\mathbf{G}^T)))^T\odot\omega\\
&+\lambda_1\sum^{d}_{j=1}4\cdot(\frac{\partial \mathcal{J}_b}{\partial \omega}\odot(\mathbf{1}_d\cdot\omega^T))^T\cdot \mathcal{J}_b \\
&+ 4\cdot\lambda_2\cdot \omega\odot \omega\odot \omega + 4\cdot\lambda_3(\sum^n_{i=1}\omega_i\odot\omega_i-1)\cdot \omega,
\end{aligned}
\label{equ:model_W_gra}
\end{equation}

where

\begin{equation}
\begin{aligned}
\mathcal{J}_b=\frac{\mathbf{X}^{T}_{.-j}\cdot(\omega\odot\omega\odot \mathbf{X}_{.j})}{(\omega\odot\omega)^T\cdot \mathbf{X}_{.j}}-\frac{\mathbf{X}^{T}_{.-j}\cdot(\omega\odot\omega\odot (\mathbf{1}_n-\mathbf{X}_{.j}))}{(\omega\odot\omega)^T\cdot (\mathbf{1}_n-\mathbf{X}_{.j})},
\end{aligned}
\label{equ:model_W}
\end{equation}

\begin{equation}
\begin{aligned}
\frac{\mathcal{J}_b}{\partial \omega}&=\frac{\mathbf{X}^{T}_{.-j}\odot(\mathbf{X}_{.j}\cdot\mathbf{1}_d^T)\cdot((\omega\odot\omega)^T\cdot \mathbf{X}_{.j})}{((\omega\odot\omega)^T\cdot \mathbf{X}_{.j})^2}\\
&-\frac{\mathbf{X}^{T}_{.-j}\cdot(\omega\odot \omega\odot \mathbf{X}_{.j})^T\cdot {\mathbf{X}_{.j}}^T}{((\omega\odot\omega)^T\cdot \mathbf{X}_{.j})^2}\\
&-\frac{\mathbf{X}^{T}_{.-j}\odot((\mathbf{1}_n-\mathbf{X}_{.j})\cdot\mathbf{1}_d^T)\cdot((\omega\odot \omega)^T\cdot(\mathbf{1}_n-\mathbf{X}_{.j}))}{((\omega\odot \omega)^T\cdot (\mathbf{1}_n-\mathbf{X}_{.j}))^2}\\
&+\frac{\mathbf{X}^{T}_{.-j}\cdot(\omega\odot\omega\odot (\mathbf{1}_n-\mathbf{X}_{.j}))\cdot(\mathbf{1}_n-\mathbf{X}_{.j})^T}{((\omega\odot \omega)^T\cdot (\mathbf{1}_n-\mathbf{X}_{.j}))^2}.
\end{aligned}
\label{equ:model_W}
\end{equation}

Then we update $\omega$ using gradient descent, and finally update $\mathbf{w}^{(t)}$ at the $t$-th iteration with:
\begin{equation}
\begin{aligned}
\mathbf{w}^{(t)}=\omega^{(t)}\odot\omega^{(t)}.
\end{aligned}
\label{equ:model_W_t}
\end{equation}

We update $\mathbf{F}$, $\mathbf{G}$ and $\mathbf{w}$ iteratively until the objective function Eq.~(\ref{equ:model_la}) converges. As we can see from Eq.~(\ref{equ:model_W_gra}), the partial gradient of term $\mathcal{J}(\omega)$ with respect to $\omega$ is not only related to decorrelation term but also influenced by the weight $k$-means loss, so the learned sample weight $\mathbf{w}$ will decorrelate the features as well as benefit for clustering.

\paragraph{Complexity Analysis} The overall complexity of each iteration of DCKM is $O(Knd+nd^2)$, which is linear with respect to $n$.

\section{Experiments}
\paragraph{Dataset}
\begin{itemize}

    \item \textbf{Office-Caltech dataset}~\cite{gong2012geodesic}. The office-caltech dataset is a collection of images from four domains (DSLR, Amazon, Webcam, Caltech), which on average have almost a thousand labeled images with 10 categories. It has been widely used in the area of transfer learning~\cite{long2014transfer}, due to the biases created from different data collecting process. We use SURF~\cite{bay2006surf} and Bag-of-Words as image features, where the dimension is 500.
    \item
    \textbf{Office-Home dataset}~\cite{venkateswara2017Deep}. It is an object recognition dataset which contains hundreds of object categories found typically in Office and Home settings. To extensively evaluate our method, we randomly sample 3 subsets from the dataset where each subset contains 10 classes (marked as OH1, OH2, OH3) and each class has hundreds of images. We also use SURF and Bag-of-Words as image features, where the dimension is 500.
\end{itemize}

\paragraph{Baselines}
Because our proposed model is based on $k$-means, $k$-means is the most direct baseline. Moreover, unsupervised feature selection algorithms can delete useless features by an unsupervised way, so we also compare with several unsupervised feature selection algorithms: RUFS~\cite{qian2013robust},  FSASL~\cite{du2015unsupervised}, and REFS~\cite{li2017reconstruction}.
All the unsupervised feature selection methods first select the useful features and then feed the selected features into the $k$-means algorithm. Furthermore, we implement three straight-forward two-step decorrelated methods to validate the necessary of jointly training.
\begin{itemize}

    \item \textbf{PCA+KM}~\cite{ding2004k}: We first perform PCA to reduce the feature dimension while removing the feature correlations, and then perform $k$-means.
    \item \textbf{Drop+KM}: We first compute each feature's correlation with other features and then drop the highly correlated features. $K$-means performs on the remaining features.
    \item \textbf{Dec+KM}: We first perform decorrelation regularizer Eq.~(\ref{equ:Balance_all}) only to learn the sample weights and then apply the weighted $k$-means.
\end{itemize}
Note that, because our model is based on $k$-means method, we mainly select $k$-means based methods as baselines to validate the effectiveness of the proposed decorrelation regularization. The decorrelation regularizer can also be easily extended to other clustering paradigms, such as the autoencoder-based clustering, which is the future work.

\paragraph{Parameter Setting and Metrics.}For DCKM, we fix $\lambda_3=1$ and select $\lambda_1$ and $\lambda_2$ from $\{10^{-2}, 10^{-1}, 1, 10, 10^2, 10^3\}$. For Drop+KM, we set the highly correlation features threshold as 0.7. For PCA+KM, following~\cite{ding2004k}, we set the reduced dimension as $K$-1, where $K$ is the number of clusters. Because all the unsupervised feature selection methods are relatively sensitive to the number of selected features, we ``grid-search'' the number of selected features from $\{50, 100, \cdots, 450\}$. And for all the methods, the number of clusters, i.e., $K$, is decided by the classes of each subdatasets. Since all the clustering algorithms depend on the initializations, we repeat all the methods 20 times using random initialization and report the average performance. We employ two widely used clustering metrics: NMI and ARI~\cite{fan2020one2multi}.

\begin{table*}[ht]
\scalebox{0.86}{

{
\begin{tabular}{|c|c||c||c|c|c||c|c|c||c||c||c|}
\hline
\multicolumn{2}{|c||}{Dataset} & Metric                                        & REFS   & FSASL  & RUFS    & PCA+KM & Drop+KM & Dec+KM & $k$-means & DCKM & Impro.  \\ \hline\hline

\multirow{8}{*}{Office-Caltech} &  \multirow{2}{*}{Amazon}      & NMI & $0.4200^*$ & 0.3948  & 0.3843   &0.3841    & 0.3529      & 0.3308      & 0.4149  & \textbf{0.4545} & 8.2\%\\ \cline{3-12}
                                &                              & ARI & $0.2248^*$ & 0.1731  & 0.1626       & 0.1647 & 0.1364      & 0.2021      & 0.1883  & \textbf{0.274} &21.9\% \\ \cline{2-12}
                                & \multirow{2}{*}{Webcam}      & NMI & $0.3904^*$ & 0.3408 & 0.3229    &0.302    & 0.3333      & 0.2971      & 0.3333  & \textbf{0.4355} &11.6\% \\ \cline{3-12}
                                &                              & ARI & $0.1636^*$ & 0.1130 & 0.0945       & 0.062 & 0.1007      & 0.1404      & 0.1007  & \textbf{0.243}&48.5\% \\
                            \cline{2-12}
                                & \multirow{2}{*}{Caltech}      & NMI & $0.2152^*$ & 0.1870& 0.1850  &0.1774     & 0.1926      & 0.1810     & 0.1778  & \textbf{0.2456} & 14.1\%\\ \cline{3-12}
                                &
                        & ARI &0.0968 & 0.0707 & 0.0715    & 0.0624  & 0.0741      & $0.0985^*$      & 0.0623  &\textbf{0.1345} & 36.5\% \\
                        \cline{2-12}
                                &
               \multirow{2}{*}{DSLR}        & NMI & $\textbf{0.4788}^*$ & 0.4774 & 0.4576  & 0.466 & 0.4526     &0.3446 & 0.4523  & 0.4739 & -1.0\%\\ \cline{3-12}
                                &                              & ARI & $0.2086^*$ & 0.1938  &0.1646       & 0.1755 & 0.1659      & 0.1736      & 0.1566  & \textbf{0.2583} & 23.8\%\\
                            \hline

\multirow{6}{*}{Office-Home} & \multirow{2}{*}{OH1}        & NMI & $0.3318^*$ & 0.3071 & 0.3124  &0.3038 & 0.2986    &0.2625 & 0.3068 & \textbf{0.3594}  & 8.3\%\\ \cline{3-12}
                                &                              & ARI & $0.1528^*$ & 0.1237  &0.1264      & 0.1223 & 0.1141      & 0.1371      & 0.1262  & \textbf{0.1926} & 26.0\%\\ \cline{2-12}
                                & \multirow{2}{*}{OH2}      & NMI & 0.3120 & $0.3126^*$  & 0.3054  &  0.3021   & 0.3042      & 0.2118      & 0.2942  & \textbf{0.3383} & 8.2\%\\ \cline{3-12}
                                &                              & ARI & $0.1504^*$ & 0.1148  & 0.1075      & 0.1097 &0.1106      & 0.1098      & 0.1035  & \textbf{0.1911} & 27.1\% \\ \cline{2-12}
                                & \multirow{2}{*}{OH3}      & NMI & $0.2220^*$ & 0.1927 & 0.1883  &0.1908     & 0.1971      & 0.1894      & 0.1922  & \textbf{0.2603} & 17.3\%\\ \cline{3-12}
                                &                              & ARI & 0.0856 & 0.0500 & 0.0517     &0.052  & 0.0529      & $0.0896^*$      & 0.0565  & \textbf{0.1330} & 48.4\%\\
                            \cline{2-12}
                            \hline
\end{tabular}}

}

\caption{Clustering results on two datasets. The `*' indicates the best performance of the baselines. Best results of all methods are indicated in bold. The last column indicates the percentage of improvements gained by the proposed method compared to the best baseline.}
\label{tab:clustering}
\end{table*}
\paragraph{Clustering Result Analysis}
Table~\ref{tab:clustering} shows the clustering results, and we have following observations. (1) Our DCKM model achieves the best performance on almost all the datasets (from 8.2\% to 48.5\% improvements compared to the best baseline). Particularly, compared with $k$-means, DCKM significantly outperforms it with the 25.1\% average improvement ratio on NMI. This well demonstrates the effectiveness of integrating the decorrelation regularizer with $k$-means. (2) Two-step decorrelated approaches (PCA+KM, Drop+KM, and Dec+KM) are not always better than $k$-means, which indicates that removing correlations between features do not necessarily benefit for clustering. We should remove the unexpected correlations which hurt the clustering performance. (3) DCKM outperforms the two-step decorrelated approaches, especially the Dec+KM method, which clearly demonstrates the importance of jointly optimizing decorrelation regularizer and clustering. (4) DCKM also outperforms various unsupervised feature selection methods. The reason is that these unsupervised feature selection methods reduce the correlation by deleting some features and some meaningful features may be deleted, while our DCKM keeps all features and removes the correlations among them. Moreover, unsupervised feature selection methods are sensitive to the number of selected features~\cite{li2017reconstruction}, but our method does not have such problem.

\begin{figure}[ht]
			\centering
			\begin{subfigure}[ht]{0.23\textwidth}
				\centering
				\includegraphics[height=1.2in, width=1.6in]{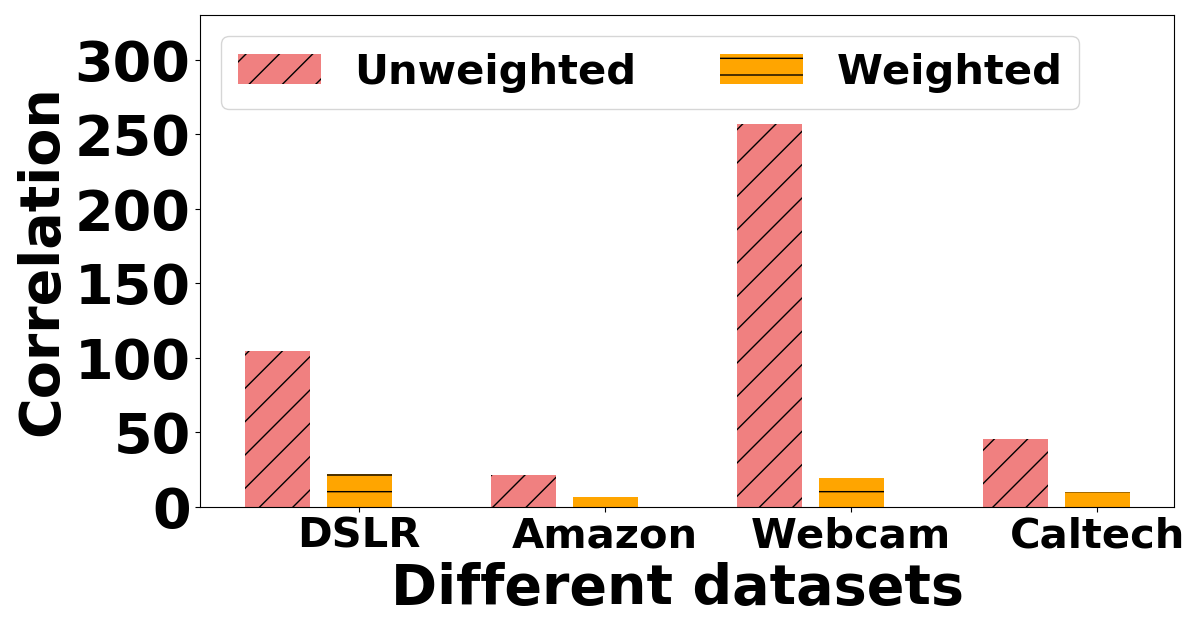}
				\caption{Office-Caltech.}
				\label{fig:add_view}
			\end{subfigure}
			\centering
			\begin{subfigure}[ht]{0.23\textwidth}
				\centering
				\includegraphics[height=1.2in, width=1.6in]{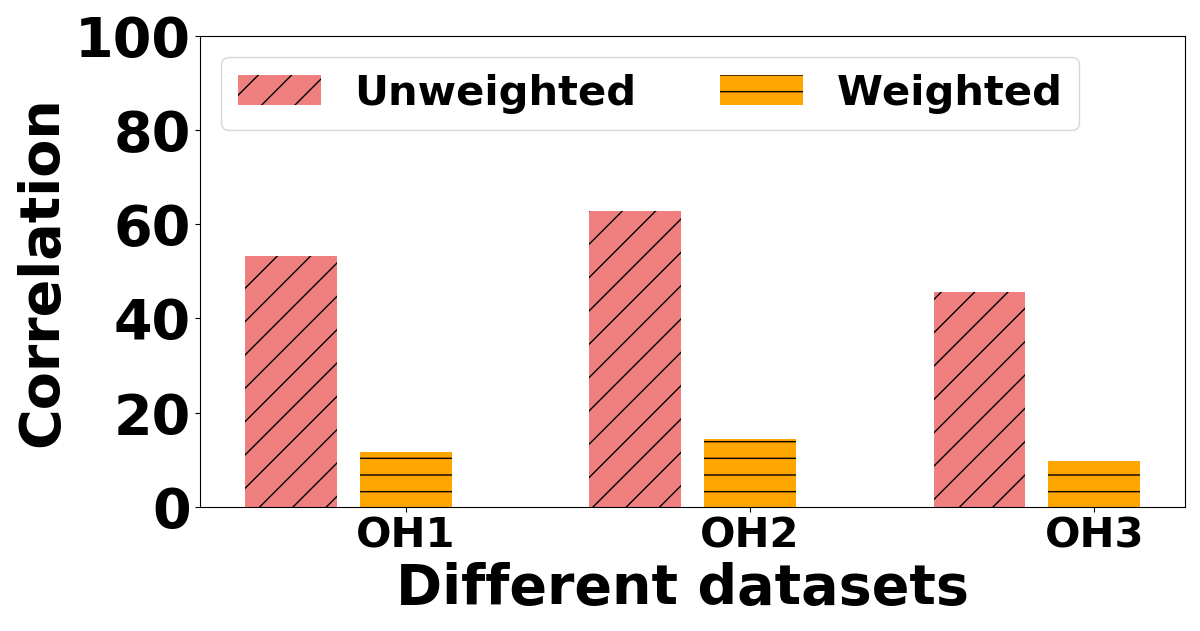}
				\caption{Office-Home.}
				\label{fig:gradient}
			\end{subfigure}
			\caption{Feature correlation analysis on unweighted and weighted datasets.}
			\label{fig:corr}
\end{figure}

\begin{figure}[ht]
			\centering
			\begin{subfigure}[ht]{0.22\textwidth}
				\centering
				\includegraphics[height=1.2in, width=1.4in]{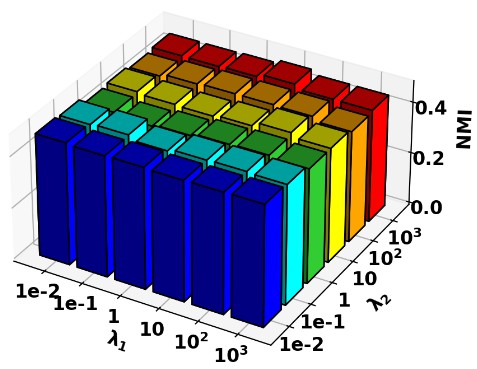}
				\caption{DSLR.}
				\label{fig:add_view}
			\end{subfigure}
			\centering
			\begin{subfigure}[ht]{0.22\textwidth}
				\centering
				\includegraphics[height=1.2in, width=1.4in]{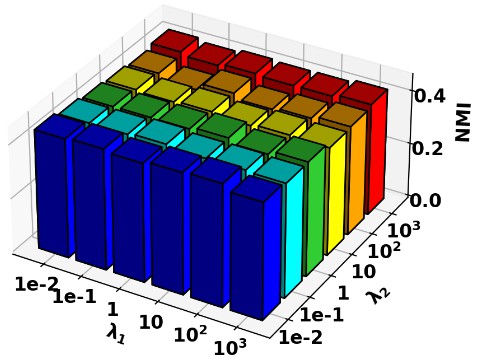}
				\caption{Amazon.}
				\label{fig:gradient}
			\end{subfigure}
			\centering
			\begin{subfigure}[ht]{0.22\textwidth}
				\centering
				\includegraphics[height=1.2in, width=1.4in]{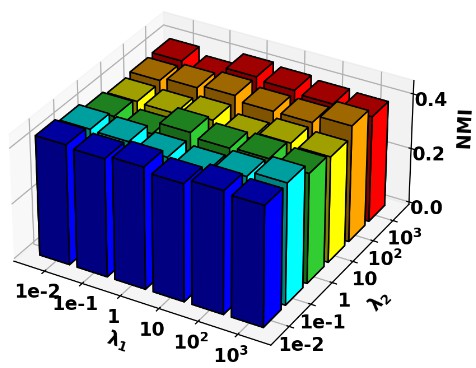}
				\caption{Webcam.}
				\label{fig:add_view}
			\end{subfigure}
			\centering
			\begin{subfigure}[ht]{0.22\textwidth}
				\centering
				\includegraphics[height=1.2in, width=1.4in]{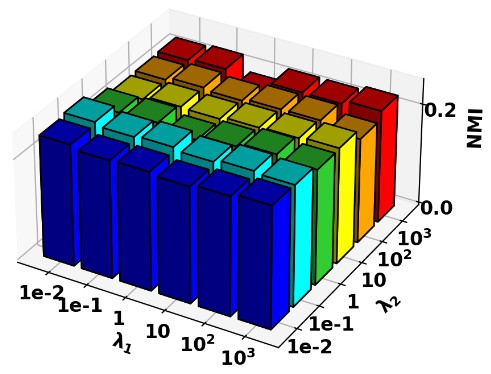}
				\caption{Caltech.}
				\label{fig:gradient}
			\end{subfigure}
			\caption{NMI of DCKM with different $\lambda_1$ and $\lambda_2$ while keeping $\lambda_3=1$ on Office-Caltech datasets.}
			\label{fig:para}
\end{figure}

\paragraph{Sample Weight Analysis} Here we analyze the effect of sample weights $\mathbf{w}$ in our model. We compute the amount of correlations in original unweighted dataset and the weighted dataset, in which the weights are the last iteration sample weights of DCKM. Following~\cite{cogswell2015reducing}, the amount of correlations is measured by the Frobenius norm of the sample cross-corvairance matrix computed from the features of samples. Figure~\ref{fig:corr} shows the amount of correlations in unweighted dataset and weighted dataset, and we can observe that the feature correlations in all the weighted datasets are reduced, demonstrating that the weights learned by DCKM can reduce the correlations between the features. Since
the major difference between DCKM and a standard $k$-means is the decorrelation regularizer, we can safely attribute
the significant improvement to the effective decorrelation regularizer and its seamless joint with $k$-means.
\paragraph{Parameters Sensitivity}
In this subsection, we study the sensitiveness of parameters. Limited by space, we just report the results on four subdatasets of Office-Caltech with $\lambda_3 = 1$ (sensitiveness under other values of $\lambda_3$ is similar) on Figure~\ref{fig:para}. The experimental results show that DCKM is relatively stable to $\lambda_1$ and $\lambda_2$ with wide ranges, indicating the robustness of DCKM.

\section{Conclusion}
In this paper, we investigate a seldom studied but important problem: clustering on data with selection bias. The data selection bias will inevitably introduce correlations between the features, making the data distribution confuse for clustering. We then propose a novel decorrelation regularized $k$-means model, which combines the feature balancing technique with $k$-means in a unified framework. Extensive experimental results well demonstrate the effectiveness of DCKM.

\section*{Acknowledgments}

This work was supported by the National Natural Science Foundation of China (No. U1936220, 61702296, 61772082, 61806020, U1936104), the National Key Research and Development Program of China (2018YFB1402600), the CCF-Tencent Open Research Fund, and the Fundamental Research Funds for the Central Universities. Kun Kuang's research was supported by the Fundamental Research Funds for the Central Universities; National Key Research and Development Program of China No. 2018AAA0101900. Shaohua Fan's research was supported by BUPT Excellent Ph.D. Students Foundation (No. CX2019127).

\clearpage
\bibliographystyle{named}
\bibliography{ijcai20.bib}

\end{document}